\documentclass[journal]{IEEEtran}
\usepackage{cite,url,subfigure,epsfig,graphicx}
\usepackage{tipa}
\usepackage{makecell}
\usepackage{multirow}
\usepackage{bbm}
\usepackage{mathrsfs}
\usepackage{array}
\usepackage{amsfonts}
\usepackage{cite,url,subfigure,epsfig,graphicx}
\usepackage{amssymb,amsmath}
\usepackage{indentfirst}
\usepackage{pifont}
\usepackage{algorithmic}
\usepackage{algorithm}
\usepackage{cases}
\usepackage{soul}
\usepackage{booktabs}
\usepackage[table]{xcolor}
\usepackage{graphicx}
\IEEEoverridecommandlockouts
\makeatletter
\usepackage{times,verbatim,amsfonts,amsmath,color}
\usepackage{hyperref}
\makeatletter
\def\UrlAlphabet{%
      \do\a\do\b\do\c\do\d\do\e\do\f\do\g\do\h\do\i\do\j%
      \do\k\do\l\do\m\do\n\do\o\do\p\do\q\do\r\do\s\do\t%
      \do\u\do\v\do\w\do\x\do\y\do\z\do\A\do\B\do\C\do\D%
      \do\E\do\F\do\G\do\H\do\I\do\J\do\K\do\L\do\M\do\N%
      \do\O\do\P\do\Q\do\R\do\S\do\T\do\U\do\V\do\W\do\X%
      \do\Y\do\Z}
\def\UrlDigits{\do\1\do\2\do\3\do\4\do\5\do\6\do\7\do\8\do\9\do\0}
\g@addto@macro{\UrlBreaks}{\UrlOrds}
\g@addto@macro{\UrlBreaks}{\UrlAlphabet}
\g@addto@macro{\UrlBreaks}{\UrlDigits}
\makeatother

\usepackage{geometry}
\usepackage[switch,pagewise]{lineno}

\begin{document}

\title{Enabling Regulatory Multi-Agent Collaboration: Architecture, Challenges, and Solutions}
\author{
\IEEEauthorblockN{
Qinnan~Hu\IEEEauthorrefmark{2}, 
Yuntao~Wang\IEEEauthorrefmark{2}, 
Yuan~Gao\IEEEauthorrefmark{2}, 
Zhou~Su\IEEEauthorrefmark{2}\IEEEauthorrefmark{1},  
Linkang~Du\IEEEauthorrefmark{2}, and 
Qichao~Xu\IEEEauthorrefmark{3}
}\\

\IEEEauthorblockA{
\IEEEauthorrefmark{2}School of Cyber Science and Engineering, Xi'an Jiaotong University, China \\
\IEEEauthorrefmark{3}School of Mechatronic Engineering and Automation, Shanghai University, China\\
\IEEEauthorrefmark{1}Corresponding author: zhousu@ieee.org 
}
}
\maketitle


\begin{abstract}
Large language models (LLMs)-empowered autonomous agents are transforming both digital and physical environments by enabling adaptive, multi-agent collaboration. While these agents offer significant opportunities across domains such as finance, healthcare, and smart manufacturing, their unpredictable behaviors and heterogeneous capabilities pose substantial governance and accountability challenges. In this paper, we propose \textcolor{black}{a hierarchical regulatory framework that combines off-chain real-time computation with on-chain Merkle-anchored auditing on a permissioned blockchain} for regulatory agent collaboration, 
\textcolor{black}{comprising an agent layer, an off-chain computation layer, and an on-chain anchoring layer.}
\textcolor{black}{This design supports sub-second latency for real-time regulatory responses while maintaining bounded on-chain cost by anchoring only compact cryptographic commitments and regulatory outcomes.} Within this framework, we design three key modules: (i) a malicious behavior forecasting module for early detection of adversarial activities, 
(ii) an agent behavior tracing and arbitration module for automated accountability, 
and (iii) a dynamic reputation evaluation module for trust assessment in collaborative scenarios.
Our approach establishes a systematic foundation for trustworthy, resilient, and scalable regulatory mechanisms in large-scale agent ecosystems. \textcolor{black}{Experimental results validate both the regulatory effectiveness and the latency efficiency of the proposed framework.} Finally, we discuss the future research directions for blockchain-enabled regulatory frameworks in multi-agent systems.
\end{abstract}


\section{Introduction}\label{sec:introduction}
\IEEEPARstart{A}{utonomous} agents are rapidly emerging as a transformative paradigm in both digital and physical environments. Unlike traditional networked devices that primarily collect and relay data, agents can independently sense, reason, and act upon their surroundings. Recent advances in large language models (LLMs) such as GPT-5 and DeepSeek amplify this trend by endowing agents with advanced reasoning, natural language interaction, and adaptive planning capabilities, enabling them to operate in increasingly complex and unstructured contexts~\cite{wang2024large}. This shift has expanded their role from passive executors of predefined commands to active decision-makers capable of adapting to dynamic conditions and interacting with other agents. As the populations of LLM-driven software agents and embodied robots grow, their large-scale collaboration is becoming critical for addressing complex, multi-faceted tasks across domains such as finance, healthcare, logistics, and smart manufacturing~\cite{liu2024heterogeneous}. 

However, such collaboration is inherently coupled with challenges of governance and accountability, 
as agent behaviors especially those empowered by LLMs are often unpredictable and difficult to regulate in real time~\cite{hu2023mo}.
These concerns highlight the need for regulatory mechanisms that ensure both operational efficiency and systemic trust in multi-agent ecosystems.
At the core of regulatory agent collaboration lie three unique characteristics:
\begin{itemize}
    \item \textit{Autonomous decision-making:} Agents operate with minimal human intervention, yet their unpredictable actions can introduce systemic risks. This necessitates mechanisms that provide auditable decision trails to ensure accountability.
    \item \textit{Social collaboration:} Agents form temporary task teams and jointly pursue goals, but limited mutual trust creates vulnerabilities, as they may exaggerate capabilities or disseminate false information during cooperation. Transparent and verifiable reputation credentials are therefore essential.
    \item \textit{Resource heterogeneity:} Agents range from virtual voice assistants to resource-constrained desktop robot pets, with diverse computational power, sensing modalities, and energy profiles. This diversity requires adaptive adversarial behavior detection for constrained environments.
\end{itemize}


Blockchain offers a promising foundation to address these regulatory challenges in multi-agent collaboration. Specifically, an immutable ledger provides auditable decision records, smart contracts enable transparent enforcement of interaction rules, and decentralized consensus ensures mutual trust without centralized authorities~\cite{karim2025ai}. By integrating blockchain, regulatory agent collaboration could achieve verifiable accountability, resilient trust management, and fair resource coordination across diverse environments. However, current blockchain solutions~\cite{9679805,10103197,10769427} remain inadequate for real-world deployments in agent ecosystems.
Specifically, they lack \textcolor{black}{(i) a low-latency trusted auditing framework, as fully on-chain designs can introduce high latency overhead and state bloat,} (ii) dynamic reputation assessment tailored to the fluid nature of inter-agent collaboration, and (iii) proactive mechanisms to forecast and detect adversarial agent behaviors.

\textcolor{black}{This paper introduces a hierarchical regulatory framework for efficient large-scale agent collaboration, consisting of three tiers: 1) the agent layer for agent identities, capabilities, and interactions; 2) the off-chain computation layer for real-time tracing, arbitration, reputation evaluation, and malicious behavior prediction; and 3) the on-chain anchoring layer for Merkle-based auditing on a permissioned blockchain.} Building upon this architecture, we design three key modules: (i) a malicious behavior forecasting module based on a diffusion model that provides early warnings of potential adversarial activities, (ii) an agent behavior tracing and arbitration module based on chaincode that enables automated accountability and dispute resolution, and (iii) a reputation evaluation module that dynamically assesses trustworthiness in collaborative scenarios.

The remainder of this paper is organized as follows. Section~\ref{background} introduces the background and unique characteristics of agent collaboration, and highlights the core challenges in large-scale multi-agent governance. Section~\ref{opportunities} explores the opportunities for blockchain-enabled regulatory frameworks for trustworthy agent collaboration. Section~\ref{case_study} provides a case study built on the proposed architecture, including agent behavior tracing and arbitration module, reputation evaluation module, and malicious behavior forecasting module. Section~\ref{future_work} outlines potential directions for future research in regulatory agent collaboration. Finally, Section~\ref{conclusion} concludes the paper.

\begin{figure}[!t]
\centering \setlength{\abovecaptionskip}{-0.1cm}
  \includegraphics[width=0.8\linewidth]{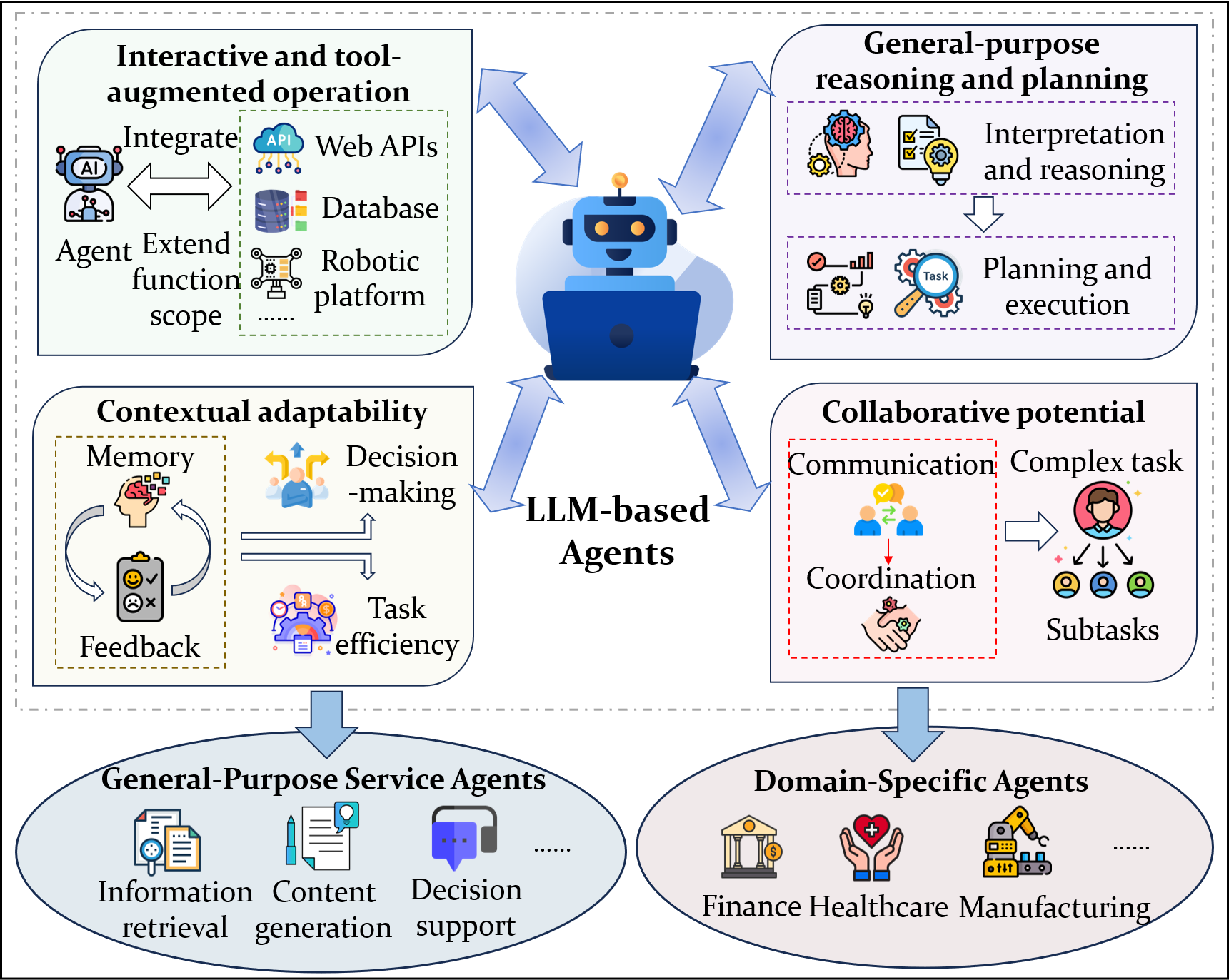}
  \caption{An overview of LLM-based agents.}\label{fig:background_agent}\vspace{-2.5mm}
\end{figure}

\section{Overview of LLM-based Agents and Key Regulation Challenges in Agent Cooperation}\label{background}

\subsection{Overview of LLM-based Agents
}
Recent advances in LLMs have catalyzed the emergence of autonomous agents that leverage the reasoning, planning, and interaction capabilities of foundation models. 
As illustrated in Fig.~\ref{fig:background_agent}, unlike static model inference, LLM-based agents can perceive tasks, decompose them into multi-step action plans, and execute them within digital and physical environments~\cite{11098567}. These agents possess several salient characteristics:
\begin{itemize}
    \item \textit{General-purpose reasoning and planning:} LLMs enable agents to adaptively interpret instructions, perform logical reasoning, and devise strategies for complex tasks.
    \item \textit{Interactive and tool-augmented operation:} Agents can interact with external systems, such as Application Programming Interfaces (APIs), databases, or robotic platforms, effectively extending their functional scope beyond text generation.
    \item \textit{Collaborative potential:} Through structured communication, agents can coordinate with other agents, distribute subtasks, and collectively achieve goals that exceed individual capacity.
\end{itemize}
These capabilities establish LLM-based agents as a transformative paradigm for intelligent automation, bridging human intent with autonomous, executable actions~\cite{wang2024large}. In practice, LLM-based agents can be broadly categorized into two groups.
\begin{itemize}
    \item \textit{General-purpose service agents} function as versatile assistants that support a wide range of tasks such as information retrieval, content generation, and decision support. These agents act as foundational utilities that bridge human intent with actionable outcomes in diverse domains.
    \item \textit{Domain-specific agents} are tailored for specialized applications, embedding expert knowledge and task-specific tools to address the requirements of particular verticals such as finance, healthcare, or scientific discovery.
\end{itemize}

\begin{figure}[!t]
\centering \setlength{\abovecaptionskip}{-0.1cm}
  \includegraphics[width=0.75\linewidth]{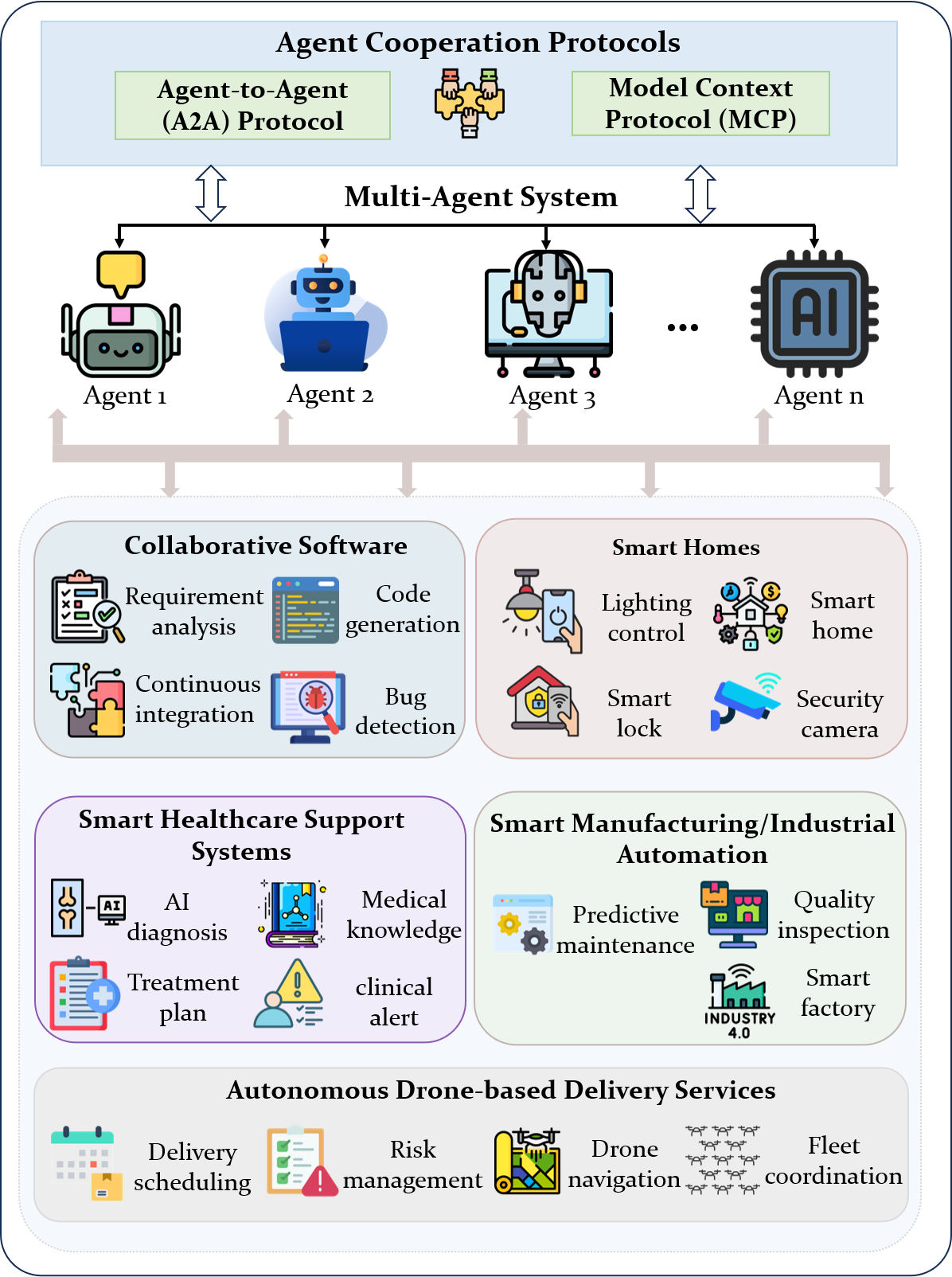}
  \caption{Applications of multi-agent cooperation.}\label{fig:application}\vspace{-2.5mm}
\end{figure}

\subsection{Applications of Multi-agent Cooperation
}
Effective agent cooperation depends on standardized communication protocols that enable both external resource access and inter-agent coordination. Two emerging protocols are reviewed:
\begin{itemize}
    \item \textit{Model Context Protocol (MCP):} MCP acts as a universal plugin interface, allowing agents to connect seamlessly with external resources such as large models, databases, APIs, and software tools. By offering a unified access layer, MCP simplifies integration and ensures secure, reliable interoperability across heterogeneous digital ecosystems.
    \item \textit{Agent-to-Agent (A2A) Protocol:} Complementing MCP, the A2A protocol provides a common language for inter-agent communication. It supports both horizontal coordination among domain-specific agents and vertical collaboration between general-purpose and specialized agents. Through A2A, agents can negotiate roles, share intermediate results, and synchronize strategies, enabling division of labor and efficient collective problem-solving.
\end{itemize}
Built upon these communication foundations, multi-agent cooperation demonstrates transformative potential across diverse domains. As shown in Fig.~\ref{fig:application}, representative applications include:
\begin{itemize}
    \item \textit{Collaborative Software Project Development:} 
    In software engineering, multiple agents can collaboratively perform requirement analysis, code generation, bug detection, testing, and security auditing. A general-purpose coordination agent manages task allocation, while specialized agents leverage MCP-enabled tools to provide complementary expertise.


    \item \textit{Smart Manufacturing and Industrial Automation:} In manufacturing environments, specialized agents can collaboratively manage production scheduling, predictive maintenance, quality inspection, and supply chain logistics. Through A2A-based information exchange and task coordination, these agents can jointly optimize production processes and respond to operational disruptions.


\end{itemize}

\begin{figure*}[!t]
\centering
\setlength{\abovecaptionskip}{-0.1cm}
\includegraphics[width=0.88\linewidth]{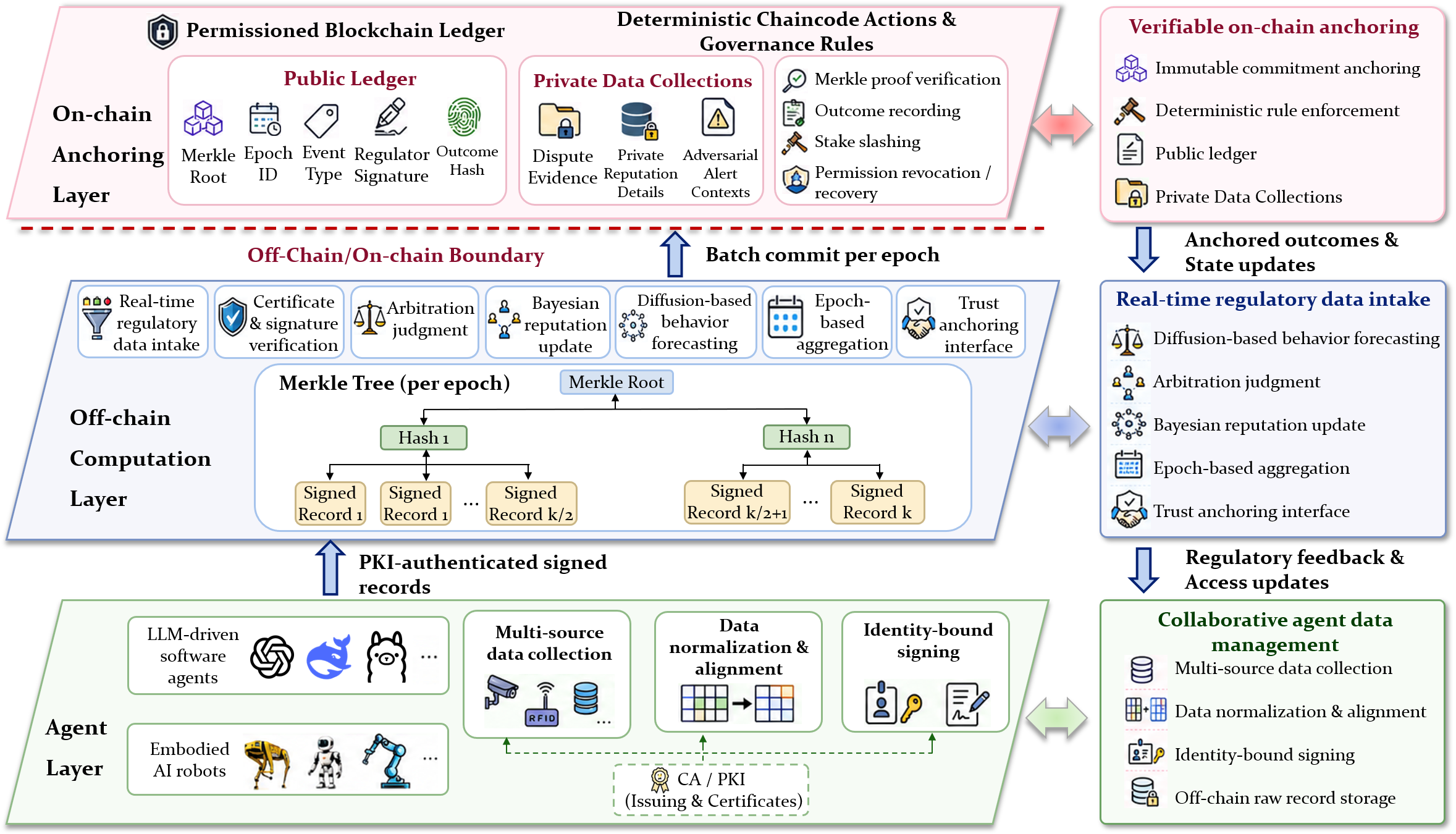}
\caption{\textcolor{black}{Hierarchical regulatory architecture with off-chain computation and on-chain anchoring. Signed agent records are aggregated into Merkle trees off-chain and periodically committed to a permissioned ledger for verifiable auditing.}}
\label{fig:framework}
\vspace{-2.5mm}
\end{figure*}

\subsection{Key Regulation Challenges in Multi-Agent Cooperation}
Ensuring trustworthy and resilient collaboration among autonomous agents presents critical regulatory challenges~\cite{xu2026trust}. Current multi-agent ecosystems face challenges in monitoring, evaluating, and preempting agent behaviors at scale, which hinder reliable cooperation in heterogeneous and decentralized environments.

\subsubsection{Lack of Proactive Adversarial Behavior Detection for Agents}
\textcolor{black}{Malicious or erroneous behaviors in LLM-based agents, such as prompt injection and  hallucination, can compromise multi-agent collaboration before their effects become visible.} 
Most existing approaches detect anomalies only after misbehavior occurs, limiting preventive action. Proactive detection frameworks that anticipate potential adversarial activities using predictive modeling or behavioral analytics are required to mitigate risks early, enabling regulators to intervene before coordination is disrupted.

\subsubsection{Lack of Automated Misbehavior Tracing and Arbitration Mechanisms for Agents}
\textcolor{black}{In LLM-based multi-agent systems, prompt injection, hallucinations, and reasoning errors can propagate across agents, making misbehavior difficult to trace and attribute.}
Existing frameworks often rely on manual auditing or centralized oversight, which is insufficient for real-time accountability. Automated tracing of agent activities and distributed arbitration mechanisms are essential to identify, resolve, and document misbehaviors without human intervention. By embedding smart contracts or algorithmic arbitration, regulatory systems can provide timely dispute resolution and maintain system integrity, even under high agent autonomy.

\subsubsection{Lack of Dynamic Reputation Assessment Mechanisms for Agents}
\textcolor{black}{The reliability of LLM-based agents can vary substantially across prompts, contexts, tool interactions, and collaborative tasks due to their context-dependent reasoning and generation behaviors.}
Existing reputation systems are primarily designed for human users or IoT devices and often fail to capture the dynamic variations in agent capabilities, behaviors, and task contexts. Therefore, developing mechanisms for continuous, context-aware reputation evaluation is essential to ensure that trust assessments accurately reflect real-time performance, promote honest reporting, and support the formation of reliable agent coalitions.


\section{Blockchain-Empowered Hierarchical Architecture for Regulatory Multi-Agent Cooperation}\label{opportunities}


\subsection{Architecture of \textcolor{black}{Hybrid On-chain/Off-chain Multi-Agent Regulation}}

As illustrated in Fig.~\ref{fig:framework}, the proposed architecture provides a \textcolor{black}{hierarchical regulatory framework that combines off-chain real-time computation with on-chain Merkle-anchored auditing on a permissioned blockchain}. It consists of three layers: \textcolor{black}{1) the agent layer, 2) the off-chain computation layer, and 3) the on-chain anchoring layer. This design avoids placing high-frequency agent operations directly on-chain, while preserving verifiable accountability, trustworthy reputation management, and auditable adversarial warnings.}

\textit{1) Agent layer:} 
The bottom layer manages heterogeneous agents, including both LLM-driven software processes and embodied robotic platforms. It provides unified mechanisms to collect, normalize, sign, and submit diverse agent-generated data.
\begin{itemize}
    \item \textit{Multi-source data collection:} The agent layer captures heterogeneous agent outputs, including low-level operational traces (\textit{e.g.,} decision inputs, sensor readings, and action logs), mid-level interaction metadata (\textit{e.g.,} task assignments and cooperation outcomes), and high-level semantic behaviors (\textit{e.g.,} coalition formation and cross-domain task execution).
    
    
    \item \textit{\textcolor{black}{Identity-bound signing:}} Each agent is bound to a verifiable identity through certificates issued by a certificate authority (CA). All submitted records are signed with the agent's private key before being sent to the regulator, enabling source authentication, non-repudiation, and resistance to forged behavioral reports. \textcolor{black}{Although signatures cannot guarantee data truthfulness, they provide accountable provenance for tracing suspicious or inconsistent records to responsible agents.}
    
    \item \textit{\textcolor{black}{Verifiable record anchoring:}} \textcolor{black}{Raw decision footprints and operational records are maintained in the regulator-side off-chain database. Instead of uploading raw data to the blockchain, the agent layer forwards signed records to the off-chain computation layer, where they are later aggregated into Merkle commitments for on-chain anchoring.}
\end{itemize}

\textit{\textcolor{black}{2) Off-chain computation layer:}} 
\textcolor{black}{The middle layer is responsible for real-time regulatory computation. It receives signed agent records, verifies their identities, executes arbitration logic, updates reputation states, and performs malicious behavior forecasting. This layer can be operated by authorized regulators or a consortium of authorized regulatory organizations. It serves as the computational core of the framework, while the blockchain only anchors compact commitments and critical outcomes.}
\begin{itemize}
    \item \textit{Real-time regulatory data intake:} 
    Signed records are continuously received from agents and buffered in arrival order for epoch-based regulatory processing. Malformed and expired records are discarded.

    
    
    

    \item \textit{\textcolor{black}{Epoch-based commitment and anchoring:}} 
    Agent records are grouped into epochs, with each epoch closing upon reaching 128 records. After epoch closure, the certificates and digital signatures of the buffered records are verified. The verified records are then aggregated into a Merkle tree, where leaf nodes correspond to signed agent records and the root summarizes the batch. The Merkle root, epoch ID, and regulatory outcomes are subsequently committed to the blockchain ledger, thereby synchronizing regulatory states with on-chain records.
    
\end{itemize}

\textit{\textcolor{black}{3) On-chain anchoring layer:}} 
\textcolor{black}{The top layer is built on a permissioned blockchain to provide tamper-resistant anchoring. 
Instead of executing computationally intensive predictive inference on-chain, this layer only records compact commitments, verifies regulatory outcomes, and enforces predefined governance rules.}

\begin{itemize}
    \item \textit{\textcolor{black}{Immutable record anchoring:}} \textcolor{black}{The ledger stores only compact commitments and key metadata, including Merkle roots, epoch IDs, and regulatory outcomes. Raw records remain off-chain, thereby reducing storage overhead and avoiding excessive on-chain state growth.}
    
    \item \textit{\textcolor{black}{Decentralized rule enforcement:}} \textcolor{black}{Chaincode\footnote{\url{https://github.com/hyperledger/fabric-chaincode-node}, accessed Sep. 10, 2026.} is used for deterministic verification and rule enforcement. Specifically, it verifies Merkle proofs, records arbitration outcomes, executes stake slashing, and revokes agent permissions according to predefined governance rules.}
    
    \item \textcolor{black}{\textit{Permissioned data access:}}
    \textcolor{black}{Sensitive regulatory data is managed through a permissioned data-availability strategy. Large-volume raw records remain in the regulator-side off-chain database, while sensitive evidence shared among authorized regulatory nodes, such as dispute evidence, private reputation details, and adversarial alert contexts, is maintained in permissioned private data spaces, \textit{e.g.,} Fabric private data collections. The public ledger retains only corresponding Merkle commitments, non-sensitive metadata, and minimized regulatory outcomes, enabling verifiable auditing without exposing private regulatory data.}
\end{itemize}

\section{Case Study: Solutions under the Proposed Architecture}\label{case_study}
\begin{figure*}[!t]
\centering \setlength{\abovecaptionskip}{-0.1cm}
  \includegraphics[width=0.95\linewidth]{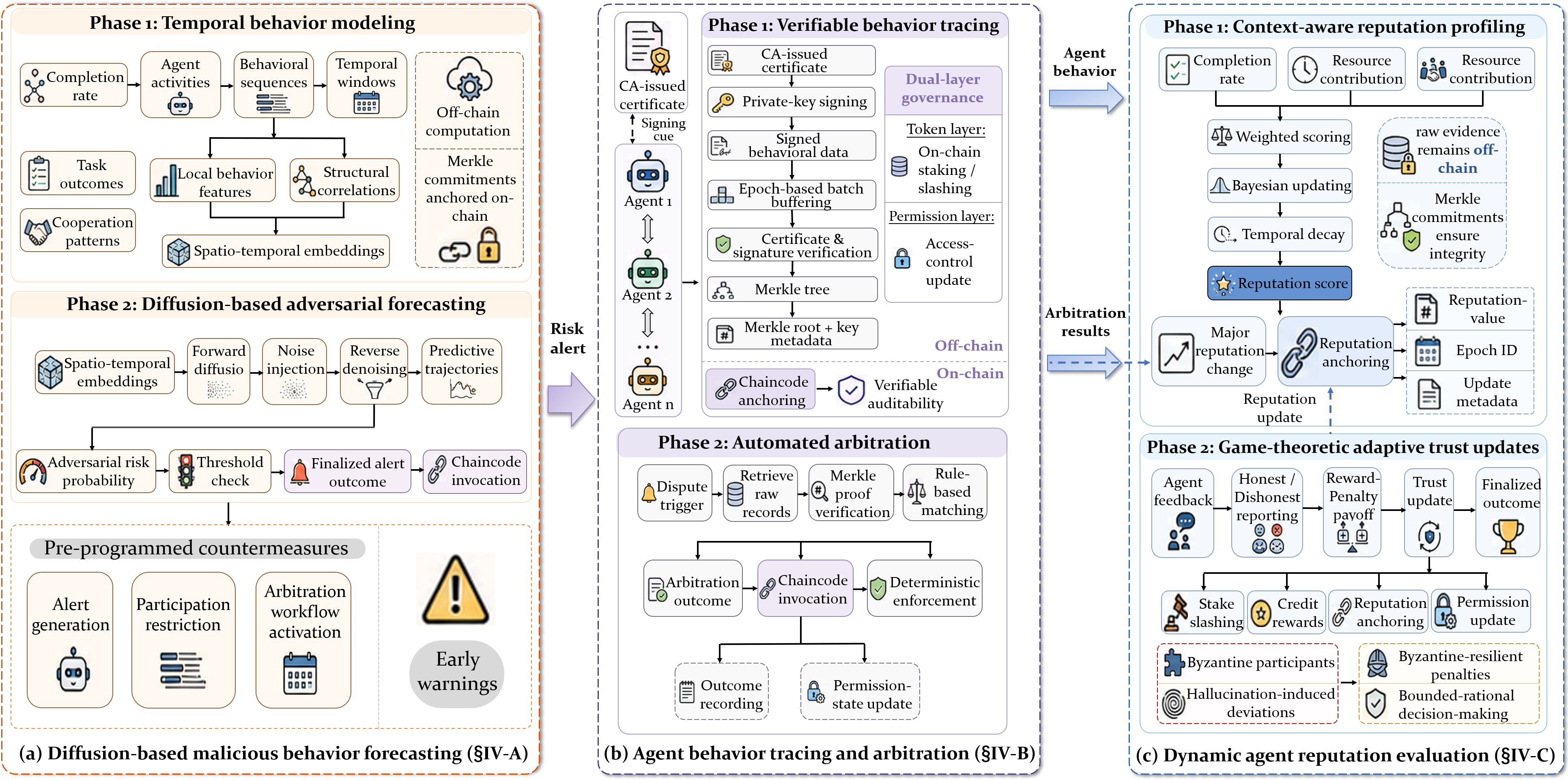}
  \caption{\textcolor{black}{Illustration of regulatory solutions under the proposed hybrid on-chain/off-chain architecture, including: (a) diffusion-based malicious behavior forecasting, (b) agent behavior tracing and arbitration, and (c) dynamic agent reputation evaluation.}}
  \label{fig:method}\vspace{-2.5mm}
\end{figure*}

Building upon the proposed blockchain-empowered regulatory architecture, we design three solutions for \textcolor{black}{multi-agent system (MAS)} ecosystems, \textit{i.e.}, diffusion-based malicious behavior forecasting, agent behavior tracing and arbitration, and dynamic agent reputation evaluation, as illustrated in Fig.~\ref{fig:method}. These solutions address proactive warnings, accountability, and trust management, respectively. We further conduct experiments to validate their effectiveness under multi-agent collaboration scenarios.

\subsection{Diffusion-Based Malicious Behavior Forecasting}

In LLM-based multi-agent ecosystems, malicious behaviors such as prompt injection and hallucination often manifest gradually before causing visible disruptions. Traditional anomaly detection approaches typically react only after such behaviors have already impacted collaboration, leading to delayed mitigation and systemic risks. To enable proactive defense, we design a diffusion-based malicious behavior forecasting module under the proposed hybrid architecture, as depicted in Fig.~\ref{fig:method}(a).

\textit{Phase 1: Temporal behavior modeling.} 
\textcolor{black}{Within the off-chain computation layer,}
\textcolor{black}{LLM-agent activities are continuously monitored and represented as multi-dimensional behavioral sequences, incorporating input prompts, generated responses, tool-use traces, inter-agent messages, and task outcomes.}
These raw sequences are first normalized and segmented into temporal windows, where local features such as tool invocation frequency, task completion rate, and agent interaction frequency are extracted to capture short-term behavioral deviations.
\textcolor{black}{For LLM-based agents, the extracted features characterize changes in tool use, task performance, and message exchanges that may be associated with prompt injection, hallucination, or tool misuse.}
To model longer-term dynamics, sliding windows and recurrent aggregation are applied to summarize periodic cooperation patterns and persistent performance trends across multiple tasks. 
Spatial correlations, such as repeated inter-agent interactions and recurring collaboration among the same agents, are further encoded to preserve structural dependencies among agents.
The resulting features are integrated into spatio-temporal embeddings that jointly reflect transient anomalies and stable behavioral trajectories. \textcolor{black}{The corresponding raw records and intermediate embeddings remain off-chain, while their integrity is preserved through epoch-based Merkle commitments anchored on-chain.}

\textit{Phase 2: Diffusion-based adversarial forecasting.} 
\textcolor{black}{Within the off-chain computation layer, a diffusion-based detection model forecasts potential malicious behaviors from the spatio-temporal embeddings.}
Specifically, behavioral trajectories are first perturbed through a forward diffusion process, where Gaussian noise is incrementally injected to simulate uncertainty and possible deviations in agent actions. During the reverse process, the detection model is trained to iteratively denoise these perturbed trajectories, gradually reconstructing the original behavioral sequence while learning the conditional probability distribution of future actions. This iterative denoising yields predictive trajectories that capture both likely cooperative behaviors and potential malicious shifts. By comparing the reconstructed trajectory with real-time observations, the system estimates the probability of malicious deviation at each step. \textcolor{black}{When the predicted risk exceeds a predefined threshold $\tau$, the finalized alert outcome is submitted to chaincode for on-chain anchoring. After verifying and recording the alert outcome, the chaincode triggers predefined countermeasures according to governance rules, including raising alerts, restricting the agent's participation in upcoming tasks, and activating the arbitration process in Sect.~\ref{sec:Arbitration}.} Unlike reactive anomaly detection, our diffusion-based forecasting mechanism provides early warnings, enabling regulators to intervene proactively before malicious agent behaviors propagate through the network.

\subsection{Agent Behavior Tracing and Arbitration}\label{sec:Arbitration}

Ensuring accountability in large-scale multi-agent ecosystems is challenging due to the unpredictability of agent actions and the difficulty of resolving disputes in real time. Traditional centralized auditing approaches are neither scalable nor resilient, as they rely on human intervention and often introduce single points of failure. To address this challenge, we design an agent behavior tracing and arbitration module under the proposed hybrid architecture. As illustrated in Fig.~\ref{fig:method}(b), its functionality is structured into two phases, \textit{i.e.,} verifiable behavior tracing and automated arbitration.

\textit{Phase 1: Verifiable behavior tracing.}
In this phase, each agent is first bound to a verifiable identity through a certificate issued by a CA. \textcolor{black}{The agent then signs its behavioral data off-chain with its private key, 
including input prompts, generated responses, tool invocation records and task outcomes.}
The signed behavioral records are buffered in arrival order and organized into epoch-based batches. After epoch closure, the agent certificates and digital signatures are verified. The verified records in each batch are then incorporated into a Merkle tree, with the corresponding Merkle root and key metadata submitted to the on-chain chaincode, enabling verifiable auditability.
The submission of agent behavioral records is automatically enforced through a dual-layer incentive mechanism.
At the token layer, staking is maintained on-chain by chaincode, and slashing is triggered by chaincode once a verified violation or missing-submission event is finalized. At the permission layer, access-control states are updated according to anchored regulatory outcomes, restricting agents that fail to provide valid records from joining subsequent collaborative tasks.

\textit{\textcolor{black}{Phase 2: Automated arbitration.}}
\textcolor{black}{Automated arbitration follows a two-stage workflow consisting of off-chain judgment and on-chain execution. 
In the off-chain stage, 
once a dispute is triggered by erroneous task outcomes or conflicting inter-agent messages,
relevant raw records are retrieved from the off-chain database, while their integrity is verified through Merkle proofs derived from the corresponding batch commitment.} 
\textcolor{black}{Rule-based matching is automatically conducted over the verified evidence to determine whether an agent has violated predefined regulatory rules, 
including task-scope violations, unauthorized tool use, response inconsistencies with task context or tool outputs, and contradictory inter-agent communications.}
\textcolor{black}{In the on-chain stage, the arbitration outcome is submitted as a transaction containing the evidence Merkle path, violation-type code, penalty-action enumeration, and signature. The on-chain arbitration contract, \textit{i.e.,} chaincode,  
then verifies the submitted outcome and executes corresponding enforcement actions,
including outcome recording and permission-state updates. 
}

\subsection{Dynamic Agent Reputation Evaluation}

Trust management is a fundamental requirement for reliable multi-agent collaboration. 
\textcolor{black}{For LLM-based agents, reliability is highly context-dependent and can vary across collaboration contexts. 
Hallucinations and tool-use errors can further cause dynamic variations in agent reliability.}
To address these challenges, we design a dynamic reputation evaluation module that combines off-chain context-aware reputation computation with on-chain anchoring and deterministic incentive enforcement, as illustrated in Fig.~\ref{fig:method}(c).


\textit{Phase 1: Context-aware reputation profiling.}
\textcolor{black}{In this phase, reputation profiling is performed within the off-chain computation layer through a multi-dimensional profiling scheme.} 
Specifically, each agent's reputation score is evaluated based on task-oriented  behavioral features, including completion rate, timeliness, resource contribution, and consistency of submitted records, which are extracted from the off-chain database and aggregated through a weighted scoring model. Bayesian updating is then applied to incorporate prior reliability estimates with newly observed evidence, while temporal decay emphasizes recent behaviors and discounts outdated records.
\textcolor{black}{The raw evidence and intermediate computation results remain off-chain, while their integrity is guaranteed by the corresponding Merkle commitments.}
\textcolor{black}{To avoid frequent on-chain updates, reputation anchoring is triggered only by major reputation changes associated with arbitration or adversarial alerts, enabling verifiable reputation auditing with reduced on-chain overhead.}

\textit{Phase 2: Game-theoretic adaptive trust updates.}
To maintain fairness and prevent dishonest reporting, the reputation update process is formulated as a repeated game among agents. In each task cycle, participating agents provide feedback on another agent's performance. \textcolor{black}{The strategy space includes \textit{honest reporting}, where agents provide truthful evaluations, and \textit{dishonest reporting}, where agents collude, manipulate feedback for strategic gain, or falsify self-reported evidence such as sensor readings, output logs, and behavioral claims.} To encourage honesty, a payoff matrix based on combined reward--penalty mechanisms is adopted. Specifically, dishonest reporting may provide short-term collusion benefits, but chaincode enforces penalty rules once such behaviors are detected, including stake slashing to impose direct stake loss, reputation degradation to reduce agent trustworthiness, and exclusion from future coalitions to restrict participation opportunities. By contrast, honest reporting is rewarded through positive reputation updates, preferential access to future coalition opportunities, increased allocation of cooperative tasks, and credit-based incentives. Consistent long-term honesty further strengthens cumulative trust records, thereby amplifying cooperative benefits. 
\textcolor{black}{To account for bounded-rational participants, such as agents with hallucination-induced deviations}, incomplete-context decisions, and misaligned objectives, as well as Byzantine agents with disruption-oriented behaviors, we incorporate the bounded-rational decision-making mechanism~\cite{jin2024bounded} into the adaptive trust update process. Specifically, beyond reward--penalty incentives, the reputation score is also adjusted according to observed task performance, submitted-record consistency, and recent behavioral evidence, which improves the robustness of trust updates against unreliable and collusive behaviors.

\subsection{Experimental Evaluation}
\textcolor{black}{We deploy a permissioned blockchain based on Hyperledger Fabric v2.5.5  with five orderer nodes running Raft consensus, and implement the regulatory logic as Go-based chaincode.}
\textcolor{black}{A server with an Intel(R) Xeon(R) Platinum 8378C CPU, 512G RAM, and dual Nvidia RTX 4090 GPUs is used for off-chain computation. The off-chain computation layer is implemented in Go for regulatory data intake and anchoring coordination, with a LevelDB database and a gRPC interface exposed to agents. The Merkle batch size is set to 128 records. Agent identities are issued by a Fabric CA, and every behavioral record is signed with the agent's ECDSA secp256r1 private key prior to submission.}
For the diffusion-based module, we employ DDPM~\cite{ho2020denoising} with an Attention U-Net backbone, configured with 1000 diffusion steps, a cosine noise schedule where beta values range from 0.0001 to 0.02, and a \textcolor{black}{risk threshold $\tau$ of 0.7.} \footnote{\textcolor{black}{A lower $\tau$ improves sensitivity to subtle behavioral deviations but increases false positives, whereas a higher $\tau$ reduces false positives but decreases sensitivity to weak deviations.}} 
\textcolor{black}{We use the TAMAS benchmark~\cite{kavathekar2025tamas} to validate the regulatory performance, which covers adversarial risks in multi-agent LLM systems across diverse attack types and real-world collaborative tasks.}
The multi-agent collaboration environment is implemented based on AutoGen,
with Qwen3-8B used as the base LLM for all agents\footnote{\textcolor{black}{We adopt Qwen3-8B and TAMAS as representative choices for the open-weight LLM backbone and the benchmark for regulatory risks in multi-agent LLM collaboration, respectively.}}.

\begin{figure}[ht]
\fbox{\centering\setlength{\abovecaptionskip}{-0.1cm}
\includegraphics[width=7.5cm]{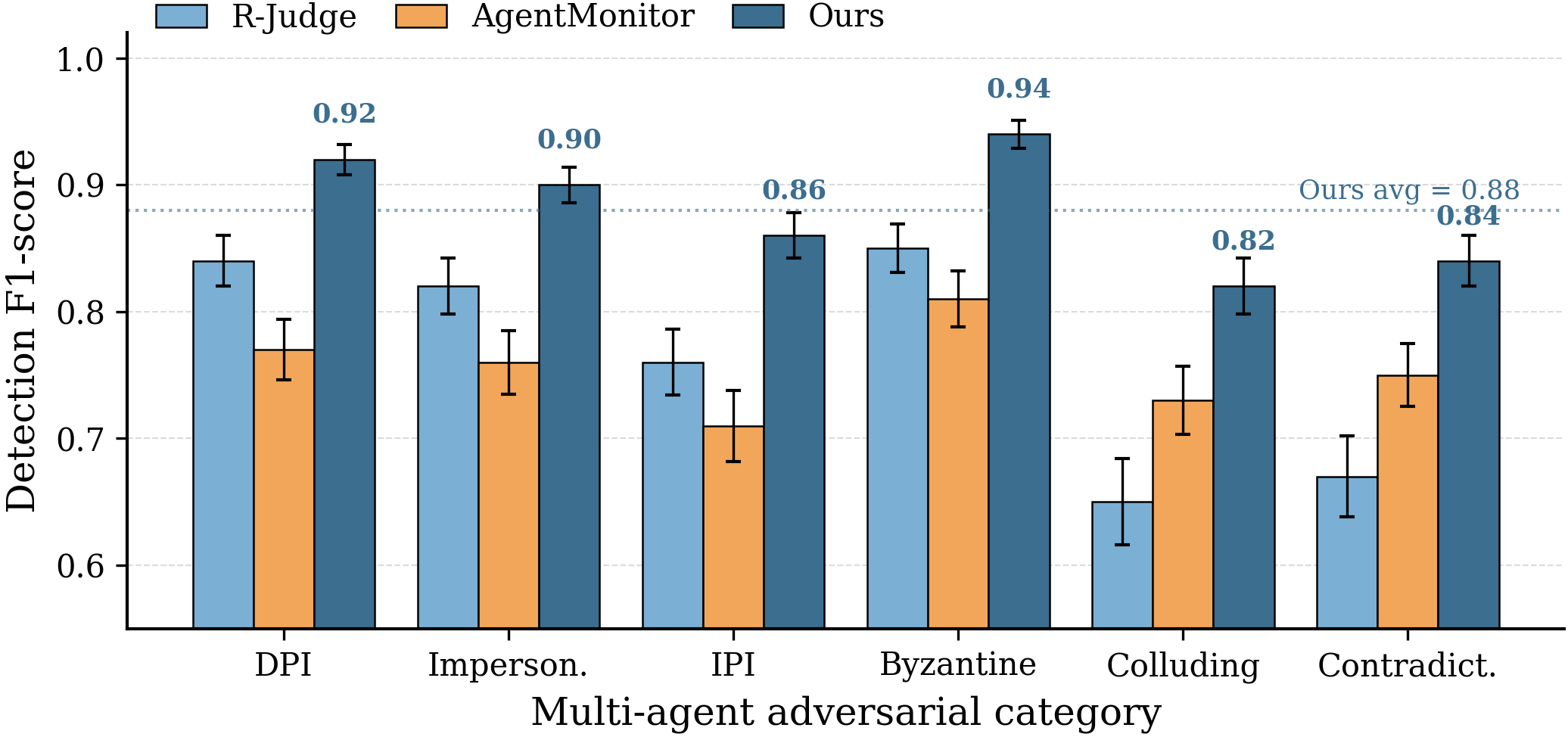}}
\caption{\textcolor{black}{Detection F1-score across the six mainstream multi-agent attack types.}}
\label{evaluation:anomaly}\vspace{-3mm}
\end{figure}

\textcolor{black}{First, Fig.~\ref{evaluation:anomaly} shows the effectiveness of our scheme in identifying malicious behaviors across six mainstream multi-agent attack categories, including Direct Prompt Injection (DPI), Impersonation, Indirect Prompt Injection (IPI), Byzantine behavior, collusion, and contradiction. Specifically, two representative regulatory detection methods, R-Judge~\cite{yuan2024rjudge} and AgentMonitor~\cite{chan2024agentmonitor}, are adopted as baselines. R-Judge is instantiated with GPT-4o as the LLM-as-a-judge, while AgentMonitor is trained under a per-scenario 60/40 train-test split.}
\textcolor{black}{As shown in Fig.~\ref{evaluation:anomaly}, our scheme consistently achieves the highest detection F1-score across the six multi-agent attack types, with average \textcolor{black}{relative improvements} of 15.0\% over R-Judge and 16.6\% over AgentMonitor. These improvements in detection performance demonstrate the effectiveness of our proposed scheme in supporting the regulation of malicious agent behaviors and capturing cross-agent adversarial dependencies.} 

\begin{figure}[ht]
\fbox{\centering\setlength{\abovecaptionskip}{-0.1cm}
\includegraphics[width=7.5cm]{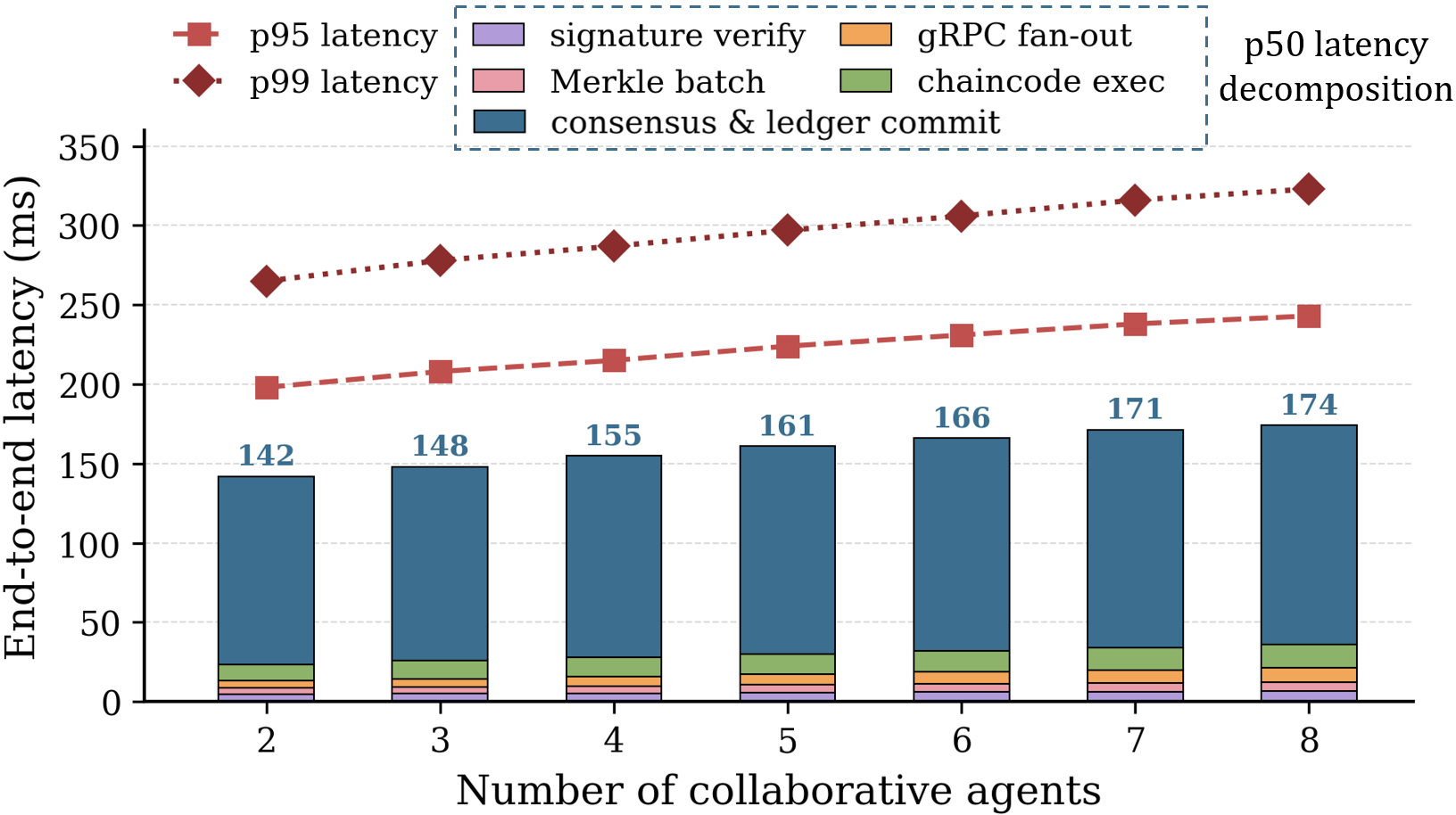}}
\caption{\textcolor{black}{End-to-end latency under increasing numbers of collaborative agents. 
}}
\label{evaluation:latency}\vspace{-2mm}
\end{figure}

\textcolor{black}{Second, Fig.~\ref{evaluation:latency} shows the latency of our scheme as the number of collaborative agents increases. 
\textcolor{black}{The end-to-end latency is measured from the closure of a full 128-record batch to the finalization of its anchor transaction on-chain, excluding batch-filling delay. Epochs close upon batch fullness.}
We measure 1k requests and report p50, p95, and p99 latency.
As shown in Fig.~\ref{evaluation:latency}, the end-to-end latency exhibits a moderate increase as the number of collaborative agents grows. 
The maximum p50 latency is below 174~ms, while the p99 latency remains below 323~ms, demonstrating that the proposed hybrid on-chain/off-chain design supports sub-second regulatory auditing with bounded tail latency.
The latency decomposition further shows that consensus and ledger commit dominate the end-to-end latency, accounting for about 81.35\%, while chaincode execution remains lightweight.}

\section{Future Research Directions}\label{future_work}
This section discusses future research directions that need to be investigated in the regulation of multi-agent cooperation.



\subsection{Cross-chain Agent Governance Frameworks}
In multi-agent ecosystems spanning multiple blockchain platforms, achieving seamless governance and interoperability presents a major challenge. Future research needs to focus on designing cross-chain protocols that synchronize agent identities, reputations, and behavioral records across heterogeneous ledgers. Techniques such as relay chains and on-chain/off-chain hybrid coordination mechanisms may enable unified regulatory policies while preserving decentralization and scalability.

\subsection{Incentive-Aligned Agent Regulation}
Ensuring agents adhere to regulatory rules requires aligning their incentives with desired behaviors. Future studies may investigate mechanism design approaches that integrate reputation, reward, and penalty systems into multi-agent collaborations. Leveraging blockchain-based tokens or reputation scores, combined with predictive modeling of agent strategies, can promote compliant behavior, discourage adversarial actions, and sustain long-term cooperation in large-scale, decentralized agent networks.

\subsection{\textcolor{black}{Verifiable Off-chain Computation}}
\textcolor{black}{
As hybrid on-chain/off-chain regulatory architectures continue to evolve, ensuring the verifiability and trustworthiness of off-chain inference becomes increasingly important. Future research may explore verifiable off-chain computation mechanisms, such as trusted execution environments (e.g., Intel SGX and AMD SEV), zero-knowledge machine learning (zkML)~\cite{chen2024zkml}, and succinct proof systems, to enable trustworthy validation of diffusion-based forecasting and LLM-assisted reasoning while preserving scalability and low-latency regulation.}


\section{Conclusion}\label{conclusion}
LLM-empowered autonomous agents are transforming our lives, offering unprecedented capabilities for adaptive decision-making, collaborative problem solving, and dynamic task execution. However, their unpredictability and lack of inherent trust mechanisms introduce significant regulatory and governance challenges. This paper has proposed a blockchain-enabled layered architecture for multi-agent collaboration. Within this framework, we have designed key modules for agent behavior tracing and arbitration, dynamic reputation evaluation, and malicious behavior forecasting. \textcolor{black}{Experimental results based on  Hyperledger Fabric demonstrate the feasibility and effectiveness of the proposed blockchain-enabled regulatory framework for multi-agent ecosystems.}
We envision that the insights and architectural principles will guide future research on adaptive and efficient multi-agent governance across diverse domains. \textcolor{black}{Future work will further evaluate the proposed framework across additional LLM backbones and multi-agent benchmarks to assess its generalizability.}

\bibliographystyle{IEEEtran}

\bibliography{ref.bib}

\vspace{-1.5cm}
\begin{IEEEbiographynophoto}{Qinnan Hu}
is currently pursuing the Ph.D degree with the School of Cyber Science and Engineering of Xi'an Jiaotong University, China. His research interests include blockchain system security and LLM Agents.
\end{IEEEbiographynophoto}\vspace{-0cm}

\vspace{-1.5cm}
\begin{IEEEbiographynophoto}{Yuntao Wang}
is currently an Assistant Professor with the School of Cyber Science and Engineering in Xi'an Jiaotong University, China. His research interests include security and privacy in UAV networks and LLM Agents.
\end{IEEEbiographynophoto}\vspace{-0cm}

\vspace{-1.5cm}
\begin{IEEEbiographynophoto}{Yuan Gao}
is currently pursuing the Ph.D degree with the School of Cyber Science and Engineering of Xi'an Jiaotong University, China. His research interests include blockchain and IoT system.
\end{IEEEbiographynophoto}\vspace{-0cm}

\vspace{-1.5cm}
\begin{IEEEbiographynophoto}{Zhou Su}
is a professor with Xi’an Jiaotong University and his research interests include multimedia communication, wireless communication, network security and network traffic. He is an Associate Editor of {\scshape IEEE Internet of Things Journal}, and {\scshape IEEE Open Journal of the Computer Society}. He is the chair of IEEE VTS
Xi'an Chapter Section.
\end{IEEEbiographynophoto}\vspace{-0cm}

\vspace{-1.5cm}
\begin{IEEEbiographynophoto}{Linkang Du} 
is currently an Assistant Professor at the School of Cyber Science and Engineering, Xi’an Jiaotong University, Xi’an, China.
His research interests include data privacy protection and trustworthy machine learning.
\end{IEEEbiographynophoto}\vspace{-0cm}

\vspace{-1.5cm}
\begin{IEEEbiographynophoto}{Qichao Xu} 
is currently an Associate Professor with Shanghai University. His research interests are in trust and security.
\end{IEEEbiographynophoto}\vspace{-0cm}

\end{document}